\documentclass[letterpaper]{article} 
\usepackage{aaai24}  
\usepackage{times}  
\usepackage{helvet}  
\usepackage{courier}  
\usepackage[hyphens]{url}  
\usepackage{graphicx} 
\usepackage{natbib}  
\usepackage{caption} 
\usepackage{algorithm}
\usepackage{algorithmic}
\usepackage{multirow}
\usepackage{amsmath}
\usepackage{amsthm}
\usepackage[utf8]{inputenc}
\usepackage{colortbl}
\usepackage{amssymb}
\usepackage{newfloat}
\usepackage{listings}
\DeclareCaptionStyle{ruled}{labelfont=normalfont,labelsep=colon,strut=off} 
\floatstyle{ruled}
\newfloat{listing}{tb}{lst}{}
\floatname{listing}{Listing}
\title{CLIP-Gaze: Towards General Gaze Estimation via Visual-Linguistic Model}
\author{Pengwei Yin\equalcontrib, Guanzhong Zeng\equalcontrib, Jingjing Wang, Di Xie}
\affiliations{
   Hikvision Research Institute, Hangzhou, China \\
  \{yinpengwei,zengguanzhong,wangjingjing9,xiedi\}@hikvision.com
}

\usepackage{bibentry}

\begin{document}

\maketitle

\begin{abstract}
Gaze estimation methods often experience significant performance degradation when evaluated across different domains, due to the domain gap between the testing and training data. Existing methods try to address this issue using various domain generalization approaches, but with little success because of the limited diversity of gaze datasets, such as appearance, wearable, and image quality. To overcome these limitations, we propose a novel framework called CLIP-Gaze that utilizes a pre-trained vision-language model to leverage its transferable knowledge. Our framework is the first to leverage the vision-and-language cross-modality approach for gaze estimation task. Specifically, we extract gaze-relevant feature by pushing it away from gaze-irrelevant features which can be flexibly constructed via language descriptions. To learn more suitable prompts, we propose a personalized context optimization method for text prompt tuning. Furthermore, we utilize the relationship among gaze samples to refine the distribution of gaze-relevant features, thereby improving the generalization capability of the gaze estimation model. Extensive experiments demonstrate the excellent performance of CLIP-Gaze over existing methods on four cross-domain evaluations.

\end{abstract}

 	\section{Introduction}
 	
 	Gaze estimation has been widely applied for human behavior and mental analysis. High-accuracy gaze estimation can provide strong support for many applications, such as human-computer interaction\cite{8542583,8542505}, saliency prediction\cite{xu2016spatio}, augmented reality\cite{10.1145/3084363.3085029} and driver monitoring systems\cite{8326022}.
    Recently, appearance-based gaze estimation methods\cite{Zhang2015AppearancebasedGE,Krafka2016EyeTF,Cheng_2018_ECCV,Cheng2020GazeEB} achieved significant results with the development of deep learning.
 	These methods achieve promising performance in the within-domain evaluation, but suffer from dramatic degradation in the cross-domain evaluations due to the domain gap.
 	
 \begin{figure}[ht]
 	\centering
 	\includegraphics[width=0.47\textwidth]{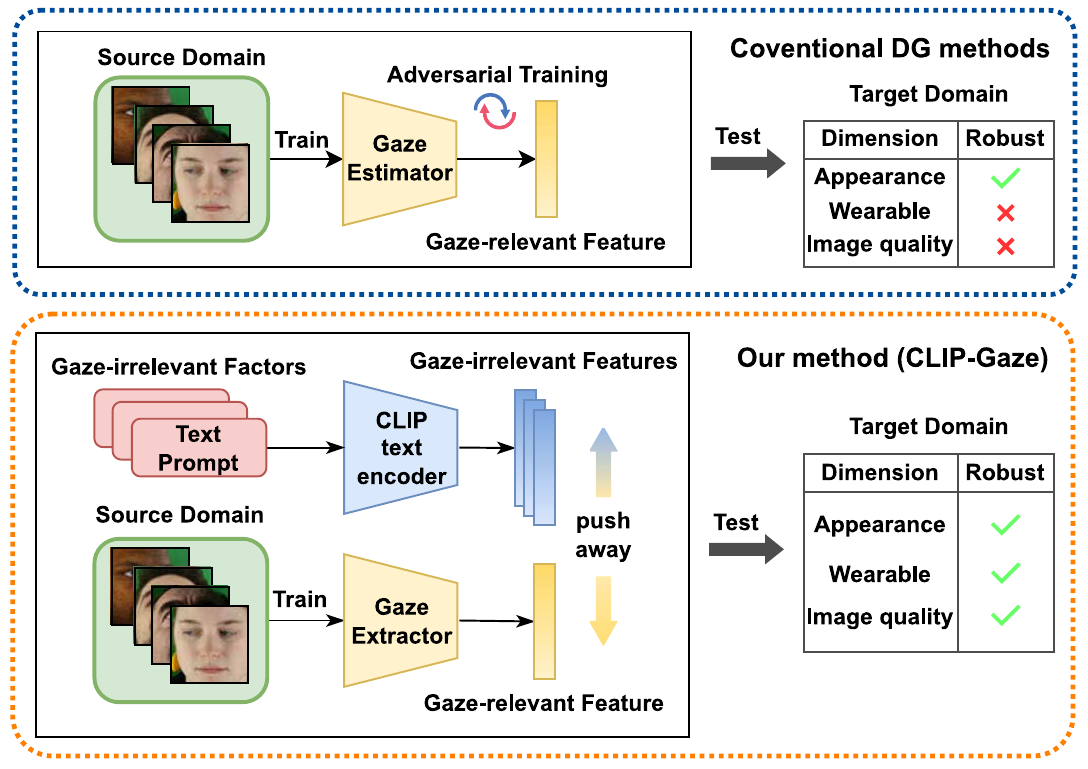}
 	\caption{(1) Top subgraph: The conventional gaze generalization approach enhances the model’s robustness by adversarial training, but can only mitigate a few gaze-irrelevant factors.
 		(2) Bottom subgraph: Our method, CLIP-Gaze, constructs a text prompt from diverse language descriptions to obtain gaze-irrelevant features, and then push away the gaze-relevant feature from gaze-irrelevant features in the feature space to handle various gaze disturbing factors and achieve a robust model.}
 	\label{tag:introduction}
 \end{figure}
 	Gaze images contain rich information, but gaze labels are mainly determined by the eye direction. The eye area only occupies a small proportion of pixels in the face image\cite{gazeconsistence}, so the model prediction results are easily affected by various disturbation factors, such as hair, beard, expressions, makeup, hat, environment illumination, sensor noise and motion blur. Thus, these gaze-irrelevant factors lead to the domain gap, and pose a great challenge for the generalization of gaze model. 	
 	To enhance the generalization of the model, some works purify the gaze feature with self-adversarial framework\cite{cheng2022puregaze} or learn stable representation with the contrastive regression loss\cite{cdg}. However, their performance is limited due to the simplicity of the source domain, which fails to cover the diverse data types of the unseen target domain. To address such a problem, common domain generalization approaches\cite{gazeconsistence,yin2022nerf} attempt to improve the diversity of the source dataset by data augmentation or data manipulation.
	However, these methods have not achieved substantial success because it is expensive to exhaustively enumerate all possible combinations of various gaze-irrelevant factors.
	Moreover, existing works have increased the data diversity in few factors such as identity, expression, and illumination, but they still struggle with many other important factors, which indicates that relying on a single visual modality is insufficient to handle all gaze-irrelevant factors.
	
	On top of this, we propose a novel method named CLIP-Gaze that leverages a pretrained vision-language model(VLM) CLIP \cite{radford2021learning} to impart general transferable knowledge to the gaze estimation model and enhance the generalization ability of extracted gaze feature.
	CLIP learns rich visual-linguistic correlations through large-scale and aligned image-text datasets, which endows it with general, powerful and flexible representation capabilities.
	Consequently, CLIP-Gaze can exploit CLIP to flexibly handle various gaze-irrelevant factors, rather than relying on expensive models or uncontrollable adversarial methods that can only deal with limited gaze-irrelevant factors.

    As shown in Fig. \ref{tag:introduction}, we first employ the CLIP text encoder to generate a set of gaze-irrelevant features as gaze distractors from flexible and diverse language descriptions, which contains all gaze-irrelevant factors mentioned above.
 	Subsequently, we maximize the distance between gaze-relevant feature and gaze-irrelevant features in the feature space via a feature separation loss function, so as to enhance the robustness of the gaze estimation model against various gaze disturbing factors.
 	Furthermore, we develop a strategy for prompt optimizing and a loss function for feature refining, i.e.,
 	Personalized Context Optimization and Feature Rank Loss to further improve the domain generality of CLIP-Gaze for gaze estimation task.
 	Personalized Context Optimization is a text prompt tuning (TPT) method that aims to avoid prompt engineering problems (a slight change in wording could make a huge difference in performance) and provide personalized text prompts for each individual. Feature Rank Loss refines the distribution of gaze-relevant features in gaze feature space by exploiting the relationship among samples, rather than only relying on the supervised loss of a single sample.

    The main contributions can be summarized as follows:
 	\begin{itemize}
 		\item We design an efficient domain-generalization framework for gaze estimation, which is the first that introduces visual-linguistic modeling into gaze estimation task, and it deals with the diverse data types not seen in the source domain through a flexible manner.
 		\item We develop a personalized text prompt tuning method to overcome prompt engineering issues and improve adaptation to the gaze estimation task. Furthermore, we propose a novel loss function based on the relationship among gaze samples to promote more reasonable feature distribution and learn robust gaze-relevant feature.
 		\item Experimental results demonstrate that our CLIP-Gaze achieves remarkable performance improvement compared with the baseline model and also outperforms the state-of-the-art domain generalization approaches on gaze estimation tasks.
 	\end{itemize}	
 	
    \begin{figure*}[htp]
        \centering
        \includegraphics[width=1\textwidth]{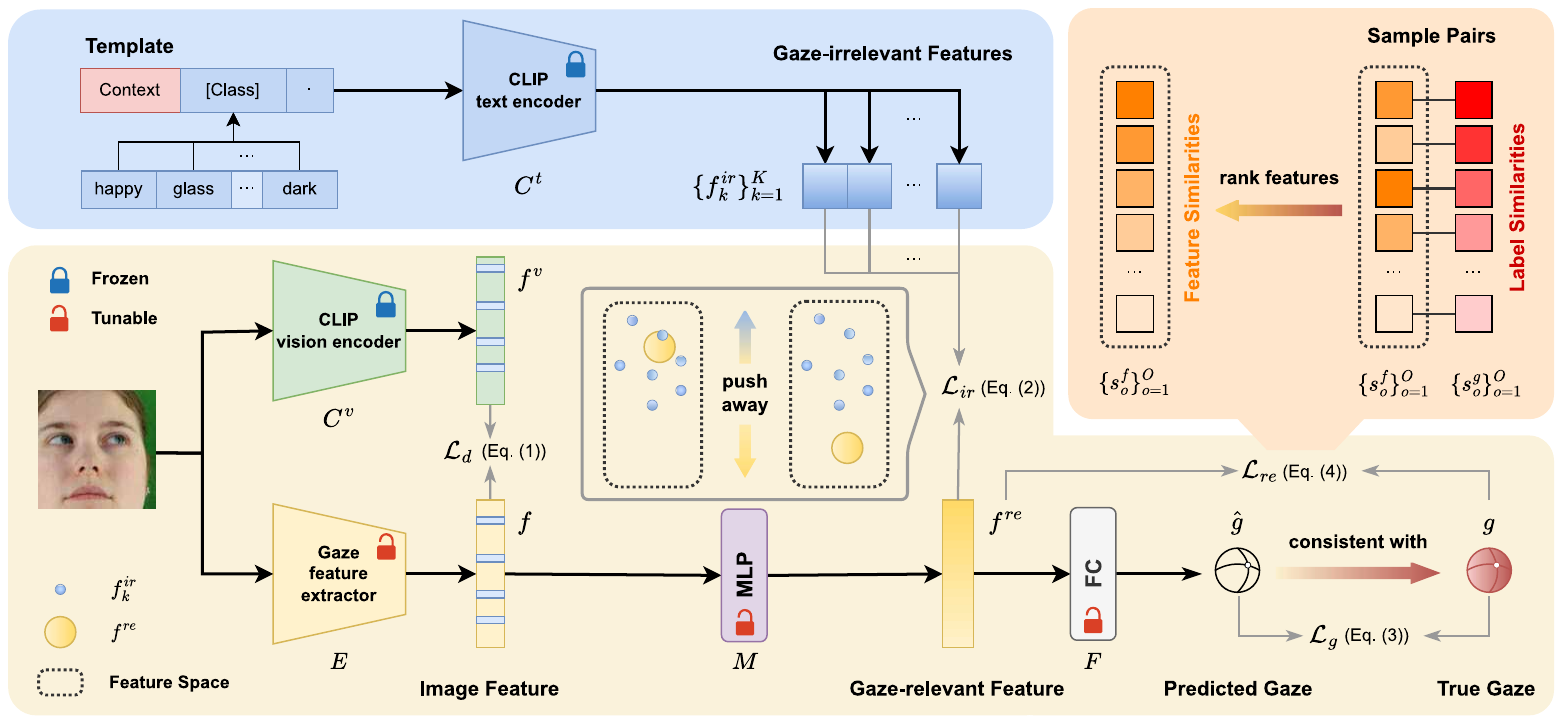}
        \caption{Overview of our CLIP-Gaze framework. We promote gaze domain generalization by introducing abundant knowledge outside the source domain to explicitly eliminate gaze-irrelevant features.}
        \label{tag:framework}
    \end{figure*}

	\section{Related Works}
	\textbf{Gaze Estimation Domain Generalization
 } Appearance-based gaze estimation has become a hotspot\cite{Zhang2015AppearancebasedGE,Krafka2016EyeTF,Cheng_2018_ECCV,Cheng2020GazeEB}, but still has many challenges on cross-domain evaluation due to the domain gap caused by various gaze-irrelevant factors.
    The common approaches typically collect diverse and extensive gaze datasets\cite{zhang2020eth,kellnhofer2019gaze360} to train a model with robust generalization capabilities, but collecting gaze data is often costly, and the diversity of the available data  remains  limited.
    This implies that we need to enhance models by using domain generalization (DG) methods, which can generalize to unseen distributions and improve cross-domain performance.
    However, most domain generalization methods are designed for classification tasks rather than regression tasks, and there are only a few studies on the generalization of gaze estimation.
    PureGaze\cite{cheng2022puregaze} proposes a self-adversarial framework to alleviate the gaze disturbation by eliminating gaze-irrelevant feature and purifing gaze-relevant feature.
    Xu $\textit{et al.}$\cite{gazeconsistence} disturb training data against gaze-irrelevant factors by using adversarial attack and data augmentation, but they neglect many other gaze-irrelevant factors that still challenge gaze estimation.

\textbf{Vision-Language Model  }
    Recently, many works have taken advantage of CLIP’s flexible text manipulation and visual alignment capabilities to enhance the open detection or generalization performance of specific tasks, such as DetCLIP\cite{NEURIPS2022_3ba96055}, DenseCLIP\cite{rao2022denseclip}, CLIP-Gap\cite{vidit2023clip}, Ordinalclip\cite{li2022ordinalclip}, CLIP-Cluster\cite{shen2023clip} and so on.
    Furthermore, to improve the performance of vision-language models on downstream tasks, a more effective approach is to learn continuous text prompts by text prompt tuning\cite{zhou2022learning,zhou2022conditional}.
    In this paper, we extend the usage of CLIP to gaze estimation, by utilizing its rich domain information to enhance the generalization ability of gaze estimation models, since it is trained on a large scale of data.

    \section{Method}
    \subsection{Preliminaries}
    A generalized gaze representation learning can be formulated as:
    \begin{equation}
        \min _{G} \mathbb{E}_{x, y} \ell(G(x, y))+\gamma \ell_{reg}
        \nonumber
    \end{equation}
    where $(x, y)$ is the input image and label, $G$ is the gaze estimator, $\gamma$ is the trade-off parameter and $\ell_{reg}$ denotes some regularization term.

    Many methods attempt to learn a generalizable gaze feature by proposing various $\ell_{reg}$, and their goals are to eliminate the influence of gaze-irrelevant factors on gaze estimation by explicit adversarial attack or data disturbation.
    However, it is difficult to cover all gaze-irrelevant factors. Although existing works have improved the robustness of models in some factors, i.e. identity, expression, and illumination, they still ignore many other important factors that challenge gaze estimation.

    Therefore, we design a flexible and scalable method for gaze-irrelevant feature construction to cover a variety of target domains. We comprehensively define the gaze-irrelevant factors from three dimensions:
 	\begin{itemize}
 		\item \textbf{Appearance.} Face images contain rich but disturbing information for gaze estimation, such as identity features (e.g., face shape, eyebrows, eyes, mouth, and nose), expression variations, texture attributes (e.g., hair and beard color and style), and other characteristics (e.g., gender and age).
 		\item \textbf{Wearable.} Besides the intrinsic factors of the face, gaze estimation is also influenced by external factors, such as wearing glasses, hats, helmets and makeup.
 		\item \textbf{Image Quality.} Except for above image content factors, sensor noise, motion blur and environment also play a role. Therefore, we define multiple types of image clarity and illuminations\cite{schlett2022face}.
 	\end{itemize}
 	
  By enumerating the gaze-irrelevant factors exhaustively, we	get a comprehensive set, which contains $K$ gaze-irrelevant factors $\{\boldsymbol{c}_k\}_{k=1}^{K}$ in total and significantly exceeds the number of factors considered by previous methods. See the supplementary material for more details.

    \subsection{CLIP-Gaze Framework}

    In this section, we will elaborate on the proposed simple and flexible gaze generalization framework named CLIP-Gaze, which leverages the available pre-trained vision-language model to introduce rich general knowledge for gaze estimation task.
    Fig. \ref{tag:framework} illustrates the whole pipeline of CLIP-Gaze that consists of two models. The first model is CLIP, which is fixed and used to generate image features $\boldsymbol{f}^{v}$ and construct multiple textual gaze-irrelevant features $\{\boldsymbol{f}^{ir}_k\}_{k=1}^K$ by defined prompt templates.

    The second model is the gaze model, which is used to extract image feature $\boldsymbol{f}$ including gaze-relevant and gaze-irrelevant features through a convolution neural network (CNN). Then we use a multi-layer perceptron (MLP) to separate gaze-relevant features $\boldsymbol{f}^{re}$ and push it away from $\{\boldsymbol{f}^{ir}_k\}_{k=1}^K$ to learning robust gaze representation. In this way, the gaze estimator can generalize to multiple target domains. Next, we will describe more details.

    \noindent
    \textbf{Construct Gaze-irrelevant Features: }
    Based on the flexibility of the VLM, we use the prompt template \texttt{"An image of a face with \{$\boldsymbol{c}_k$\}."} to construct gaze-irrelevant features $\{\boldsymbol{f}^{ir}_k\}_{k=1}^K$ about the content of input images via CLIP text encoder $C^t$, where $\boldsymbol{c}_k$ is a gaze-irrelevant factor from the set $\{\boldsymbol{c}_k\}_{k=1}^{K}$.

    \noindent
    \textbf{Distill to CLIP Feature Space: }
    For a given input image $x$, the whole gaze estimation process can be formulated as $\hat{\boldsymbol{g}} = F(M(E(x)))$. Specifically, we regard ResNet-18\cite{he2016deep} as our backbone $E$ to extract image feature $\boldsymbol{f}$, then use a MLP as feature filter $M$ to separate gaze-relevant feature $\boldsymbol{f}^{re}$, and lastly we predict the gaze direction through a fully connected (FC) layer $F$.

    The gaze-irrelevant features $\{\boldsymbol{f}_k^{ir}\}_{k=1}^{K}$ constructed above are fixed and located in the CLIP feature space.
    To remove gaze-irrelevant information from image features $\boldsymbol{f}$ and remain gaze-relevant features $\boldsymbol{f}^{re}$, we first align gaze image feature to the CLIP feature space by the following loss function:
    \begin{equation}
	\label{loss:l_distill}
            \mathcal{L}_{d}\left(\boldsymbol{f}, \boldsymbol{f}^{v}\right)=1-\left(\frac{\boldsymbol{f} \cdot \boldsymbol{f}^{v}}{\|\boldsymbol{f}\|\left\|\boldsymbol{f}^{v}\right\|}+1\right) * 0.5
        \end{equation}
    Where $\boldsymbol{f}^{v}$ is the feature extracted by CLIP vision encoder $C^v$. The loss value is normalized to the range of 0 to 1.

    \noindent
    \textbf{Separate Gaze-relevant Feature: }
    To force the feature filter $M$ to extract gaze-relevant feature $\boldsymbol{f}^{re}$, we minimize the similarity between $\boldsymbol{f}^{re}$ and $\boldsymbol{f}^{ir}_k$ where $k$ from 1 to $K$. Note that the gaze-irrelevant factors corresponding to each sample may be different, we define the $\tilde{w}_{k}$ to summarize the correlation between one sample and $k$-th irrelevant factor. $\tilde{w}_{k}$ is used to alleviate the influence of gaze-irrelevant factors that are not involved in this image.

    We use $\tilde{W}$ to represent the set of $\tilde{w}_{k}$, and $\tilde{W}= softmax(w_{1}, \dots, w_{K})$, where $w_{k}$ is the degree of correlation between current sample and $k$-th irrelevant factor and can be described as:
    \begin{equation}
        w_{k}=\frac{\boldsymbol{f} \cdot \boldsymbol{f}_k^{i r}}{\left\|\boldsymbol{f}\right\|\left\|\boldsymbol{f}_k^{i r}\right\|}
        \nonumber
    \end{equation}

    Each sample has $K$ gaze-irrelevant features elimination loss values, which will be re-weight by $\tilde{w}_{k}$ and the irrelevant loss is formally expressed as:
        \begin{equation}
	\label{loss:l_irre}
        \mathcal{L}_{ir}\left(\boldsymbol{f}^{re},\left\{\boldsymbol{f}_k^{i r}\right\}_{k=1}^K\right)=\sum_{k=1}^K \tilde{w}_k \frac{\boldsymbol{f}^{r e} \cdot \boldsymbol{f}_k^{i r}}{\left\|\boldsymbol{f}^{r e}\right\|\left\|\boldsymbol{f}_k^{i r}\right\|}
        \end{equation}

    \noindent
    \textbf{Gaze Estimation: }
    Lastly, we map the gaze-relevant feature $\boldsymbol{f}^{re}$ to gaze direction $\hat{\boldsymbol{g}}$, the gaze loss function is defined as:
    \begin{equation}
    \label{loss:l_gaze}
    \mathcal{L}_{g}\left(\hat{\boldsymbol{g}},\boldsymbol{g}\right)=\arccos \left(\frac{\hat{\boldsymbol{g}} \cdot \boldsymbol{g}}{\|\hat{\boldsymbol{g}}\| \|\boldsymbol{g}\|}\right)
    \end{equation}
where $g$ is the gaze label.

    \noindent
    \textbf{Potential Issues and Improvement Schemes: }
    Employing a suitable prompt template for CLIP-Gaze is non-trivial, as it requires prior knowledge about the gaze estimation task and proficiency in the language model’s underlying mechanism\cite{radford2021learning}.
    Hence, we propose a personalized text prompt tuning method to generate $\{\boldsymbol{f}_k^{ir}\}_{k=1}^{K}$ for each person.

    Moreover, it only learns robust features from individual samples without further exploring the relationships between samples, which is suboptimal for the regression task. According to widely accepted understanding\cite{cdg}, the label and feature relationships among samples should exhibit a strong correlation. To tackle this limitation, we propose a novel loss function that explores the relationship among gaze samples and constructs a reasonable gaze-relevant feature distribution.

        \begin{figure}[htp]
        \centering
        \includegraphics[width=0.47\textwidth]{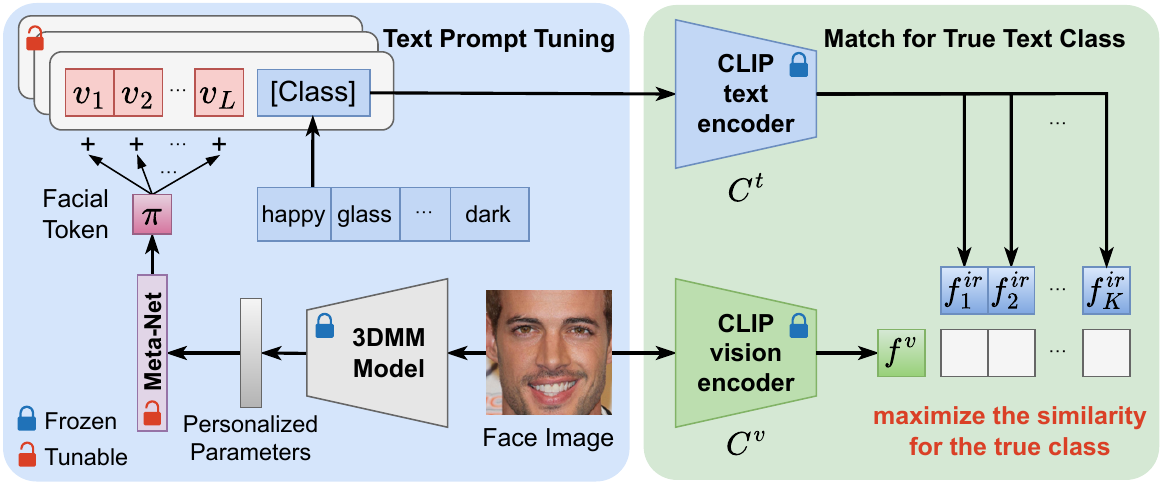}
        \caption{Our method, Personalized Context Optimization (PCO), has two learnable components: a context vector set and a lightweight neural network (Meta-Net) that produces a facial token for each identity, while the vision encoder, text encoder and 3DMM model are froze during training.}
        \label{tag:PCO}
    \end{figure}

    \subsection{Personalized Context Optimization}
    To avoid the prompt engineering issues, a common approach is to learn prompts by prompt tuning. However, existing text prompt tuning methods may not be suitable for gaze estimation.
    CoOp \cite{zhou2022learning} learns only one prompt for each class, which may fail to fully capture the gaze-irrelevant features, since each individual has a personalized facial property.
    CoCoOp\cite{zhou2022conditional} learns a prompt conditioned on each input image, by learning a lightweight neural network to impose the image content into the prompt. However, directly using the whole image content to learn the prompt may introduce some detrimental information into the prompt, since the image content also contain gaze-related information, and we want to use the prompt to extract only gaze-irrelevant feature. To this end, we propose a novel personalized context optimization (PCO)  method. Fig. \ref{tag:PCO} illustrates our approach.

    \textbf{Learn Personalizing Text Prompt: }
    First, our PCO utilizes face attribute classification as the proxy task to optimize the prompts. Specifically, we perform text prompt tuning on CelebA\cite{liu2015faceattributes} , a large-scale and diverse face attributes dataset, to obtain a more robust gaze-irrelevant features. We extend its attribute number and leverage the CLIP model to get the pseudo labels as the attribute label for prompt tuning, inspired by previous works\cite{abdelfattah2023cdul,schlett2022face}.
    See the supplementary material for more details about how to generate pseudo labels.
    We follow CoOp\cite{zhou2022learning} and introduce $L$ learnable context vectors, $\{\boldsymbol{v}_1,\boldsymbol{v}_2, \dots,\boldsymbol{v}_L\}$, each with the same dimension as the word embedding. The prompt for the $i$-th class, denoted by $\boldsymbol{t}_k$, is defined as $\{\boldsymbol{v}_1,\boldsymbol{v}_2, \dots,\boldsymbol{v}_L, \boldsymbol{c}_k\}$ where $\boldsymbol{c}_k$ is the word embedding corresponding to one class in the class name set.

    Next, to generate personalized text prompts for each individual and reduce gaze-related information, we use a pre-trained 3D Morphable Model (3DMM)\cite{tran2018nonlinear} to extract the personalization parameter $\boldsymbol{f}^m$ from the most frontal face image of each individual. Specifically, we use the 3DMM coefficients corresponding to identity as $\boldsymbol{f}^m$ to capture only the identity features of the face.
    Then, we feed $\boldsymbol{f}^m$ to Meta-Net to obtain an input-conditional token $\boldsymbol{\pi}$.
    Now, each personalizing context token is obtained by $\boldsymbol{v}_i(\boldsymbol{f}^m)=\boldsymbol{v}_i+\boldsymbol{\pi}$. The prompt for the $i$-th class is thus conditioned on the input, i.e., $\boldsymbol{t}_k(\boldsymbol{f}^m)=\{\boldsymbol{v}_1(\boldsymbol{f}^m),\boldsymbol{v}_2(\boldsymbol{f}^m), \dots, \boldsymbol{v}_L(\boldsymbol{f}^m), \boldsymbol{c}_k \}$.

    Furthermore, we feed $\boldsymbol{t}_k(\boldsymbol{f}^m)$ into text encoder $\boldsymbol{C}^t$ of CLIP to obtain text feature $\boldsymbol{f}^t_k$, and obtain face image feature   $\boldsymbol{f}^v$ via vision encoder $\boldsymbol{C}^v$ of CLIP.
    During training, we specify the negative words for each category, such as “\texttt{not happy}” for the factor “\texttt{happy}”, and  we could obtain negative text features $\overline{\boldsymbol{f}^t_k}$
    via aforementioned operations.
    Then, we could maximize the prediction probability $p_k$ for each gaze-irrelevant language description via a classification  loss function, which is formulated as:
    \begin{equation}
    p_k=\frac
    {\exp(\textrm{sim}(\boldsymbol{f}^v ,\boldsymbol{f}_k^t)/\tau)}
    {\exp(\textrm{sim}(\boldsymbol{f}^v,\boldsymbol{f}_k^t)/\tau) + \exp(\textrm{sim}(\boldsymbol{f}^v ,\overline{\boldsymbol{f}^t_k})/\tau)}
    \nonumber
    \end{equation}
    where $\tau$ is a temperature parameter learned by CLIP and $\textrm{sim}$(·,·) denotes cosine similarity.

    \textbf{Construct Personalizing Gaze-irrelevant Features: }
    After finishing text prompt tuning, we first leverage the identity labels provided by ETH-XGaze and Gaze360 to select a frontal face image for each subject, and input this image and gaze-irrelevant factors $\{\boldsymbol{c}_k\}_{k=1}^{K}$ to our PCO module to construct $K$ gaze-irrelevant features $\{\boldsymbol{f}^{ir}_k\}_{k=1}^K$ for each identity in the gaze training dataset.
	
    \subsection{Rank Gaze-Relevant Features}
    In this part, we refine the extracted gaze-relevant features through re-thinking the relationship of samples.

    Intuitively, the gaze-relevant features with similar gaze directions should be close, which is suitable for multiple gaze domains. Derived from this idea,
    a contrastive regression loss \cite{cdg} called CRLoss was proposed.
    CRLoss sets a threshold to push features with gaze angular difference larger than the threshold apart, and pull features with gaze angular difference less than the threshold together.
    However, it is hard to determine the threshold and it discards the finegrained relationships among gaze features,  since gaze estimation is a continuous regression task. To deal with this deficiency, we explore the relationship among gaze-relevant features from a novel perspective.

    We know that multiple gaze samples can be paired in pairs, and each pair of samples can be used to calculate a label similarity $s^g$ and a feature similarity $s^f$. As shown in the upper right box of Fig. \ref{tag:framework}, for multiple sample pairs, the color intensity represents the similarity level, with darker color indicating higher similarity. Then we rank all pairs from high to low according to the label similarities of each pair, and impose penalties to force the feature similarities sequence to maintain the same order as the label similarities. At last, the distribution of gaze-relevant features will be more reasonable, and the gaze feature also can be more robust.

    Specifically, based on the gaze label $\boldsymbol{g}$ and the gaze-relevant feature $\boldsymbol{f}^{re}$, a pair is composed of sample $i$ and sample $j$, we calculate a label similarity $s_{ij}^g$  and a feature similarity $s_{ij}^f$  as:

    \begin{equation}
        s_{i j}^g=\frac{\boldsymbol{g}_i \cdot \boldsymbol{g}_j}{\left\|\boldsymbol{g}_i\right\|\left\|\boldsymbol{g}_j\right\|}, \quad s_{i j}^f=\frac{\boldsymbol{f}_i^{r e} \cdot \boldsymbol{f}_j^{r e}}{\left\|\boldsymbol{f}_i^{r e}\right\|\left\|\boldsymbol{f}_j^{r e}\right\|}
        \nonumber
    \end{equation}

    For pair $p_1$ and pair $p_2$ we use  $\mathcal{L}_{re}$ to re-rank gaze-relevant features:
    \begin{equation}
	\label{loss:l_rela}
        \mathcal{L}_{re}=\max \left(0, -\mathbb{S}_{12} *\left(s_1^f-s_2^f\right)\right)
    \end{equation}
    Where $\mathbb{S}_{12}$ evaluates to 1 if $s^g_1 > s^g_2$, while evaluating to -1 when $s^g_1 < s^g_2$.

    For all samples in a mini-batch of size $B$, we construct $O=\frac{\left ( B*\left ( B-1 \right ) \right )}{2}$ pairs of samples, then randomly choose two pairs $O$ times to calculate the total rank loss.

    \subsection{Total Loss Function}
    In summary, the total loss function applied in our method is:
	\begin{equation}
		\mathcal{L} = \mathcal{L}_{g}
        + \lambda_{1}\mathcal{L}_{d}
        + \lambda_{2}\mathcal{L}_{ir}
        + \lambda_{3}\mathcal{L}_{re}
        \nonumber
	\end{equation}

    $\lambda_{1}, \lambda_{2}, \lambda_{3}$ are hyper-parameters, and we empirically set $\lambda_{1} = \lambda_{2} = \lambda_{3} = 1.0$.

	\section{Experiments}
	\subsection{Experiment Details }
	\subsubsection{Gaze Data Details}
	To verify the performance of our method in the gaze estimation task, we use ETH-XGaze\cite{zhang2020eth} and Gaze360\cite{kellnhofer2019gaze360} as training set, and test the gaze model on MPIIFaceGaze\cite{zhang2017mpiigaze} and Eye-Diap\cite{funes2014eyediap}.
    Thus, we totally evaluate on four cross-domain task, and denote them as $\mathcal{D}_\mathrm{E}$ (ETH-XGaze)$\rightarrow$$\mathcal{D}_\mathrm{M}$(MPIIFaceGaze), $\mathcal{D}_\mathrm{E}$$\rightarrow$$\mathcal{D}_\mathrm{D}$(EyeDiap), $\mathcal{D}_\mathrm{G}$(Gaze360)$\rightarrow$$\mathcal{D}_\mathrm{M}$, $\mathcal{D}_\mathrm{G}$$\rightarrow$$\mathcal{D}_\mathrm{D}$.
	See Supplementary Material for more details on the data pre-processing.
        \subsubsection{Comparison Methods}
    For Baseline, we only use $\mathcal{L}_{g}$ as training loss. For DG methods, we choose CDG\cite{cdg}, PureGaze and Xu $\textit{et al.}$’s method\cite{gazeconsistence} for comparison and use the results report by the author. Additionally, the results of SOTA UDA methods, including PnP-GA\cite{pnpga}, RUDA\cite{ruda}, CRGA\cite{cdg}, LatentGaze\cite{latentgaze}, Liu $\textit{et al.}$’s work\cite{jitter} and UnReGA\cite{unrega} as a reference.

        \subsubsection{Implementation Details}
    We conduct the experiments on a single Tesla V100 GPU. We resize and normalize all the images to $224\times224$ and [0, 1]. We set the batch size to 128 and train the model for 30 epochs on ETH-XGaze and Gaze360. See Supplementary Materials for more details on the network, training setting, CLIP setting and others.
	
    \subsection{Performance Comparison with SOTA Methods}
	\begin{table}[ht]
		\renewcommand\arraystretch{1.5}
		\setlength{\tabcolsep}{0.6mm}
		\centering
		\scalebox{1.0}{
			\begin{tabular}{|l|l|c|c|c|c|c|c|c|}
				\hline
                \multirow{2}{*}{\scalebox{0.9}{Task}}& 	\multirow{2}{*}{\scalebox{0.9}{Methods}}& \multirow{2}{*}{\scalebox{0.9}{$\mid\mathcal{D}_t \mid$}}&
                \scalebox{0.9}{$\mathcal{D}_\mathrm{E}  $} &
                \scalebox{0.9}{$\mathcal{D}_\mathrm{E}  $} &
                \scalebox{0.9}{$\mathcal{D}_\mathrm{G}  $} &
                \scalebox{0.9}{$\mathcal{D}_\mathrm{G}  $} &
                \multirow{2}{*}{Avg}
                \\
                & 	 &  &
                \scalebox{0.9}{$ \rightarrow \mathcal{D}_\mathrm{M}$} &
                \scalebox{0.9}{$ \rightarrow \mathcal{D}_\mathrm{D}$} &
                \scalebox{0.9}{$ \rightarrow \mathcal{D}_\mathrm{M}$} &
                \scalebox{0.9}{$ \rightarrow \mathcal{D}_\mathrm{D}$} &
                \\
				\hline \hline
                \multirow{6}{*}{ \scalebox{0.9}{DG} }
                & \scalebox{0.9}{Baseline} & 0 & 8.35 & 9.66 & 7.58 & 9.01 & 8.65 \\
                & \scalebox{0.9}{PureGaze} & 0 & 7.08 & $\underline{7.48}$ & 9.28 & 9.32 & 8.29 \\
                & \scalebox{0.9}{CDG} $^\ddagger$ & 0 & 6.73 & 7.95 & $\underline{7.03}$ & \underline{7.27} & $\underline{7.25}$ \\
                & \scalebox{0.9}{Xu $\textit{et al.}$} &  0 & $\underline{6.50}$ & $\textbf{7.44}$ & 7.55 & 9.03 & 7.63 \\
                & \scalebox{0.9}{CLIP-Gaze$^-$}&  0 & 7.04 & 8.51 & 7.55 & 7.73 & 7.71 \\
                & \scalebox{0.9}{CLIP-Gaze} &  0 & $\textbf{6.41}$ & 7.51 & $\textbf{6.89}$ & $\textbf{7.06}$ & $\textbf{6.97}$ \\
 			\hline \hline
				\multirow{6}{*}{\scalebox{0.9}{UDA}}
				& \scalebox{0.9}{PnP-GA ${^*}$} & 10 & 5.53 & 5.87 & 6.18 & 7.92 & 6.38 \\
				& \scalebox{0.9}{RUDA} & 100 & 5.70 & 6.29 & 6.20 & 5.86 & 6.01 \\
				& \scalebox{0.9}{CRGA} & $>0$ & 5.48 & 5.66 & 5.89 & 6.49 & 5.88 \\
				& \scalebox{0.9}{LatentGaze} & $100$ & 5.21 & 7.81 & - & - & 6.51 \\
				& \scalebox{0.9}{Liu $\textit{et al.}$} & 100 & 5.35 & 6.62 & 7.18 & 8.61 & 6.94 \\
				& \scalebox{0.9}{UnReGA} & 100 & 5.11 & 5.70 & 5.42 & 5.80 & 5.51 \\
				\hline \hline
				\multirow{2}{*}{\scalebox{0.9}{SDA}}
				& \scalebox{0.9}{Baseline $^\star$} & 100 & 4.63 & 5.86 & 5.67 & 6.26 & 5.61 \\
				& \scalebox{0.9}{CLIP-Gaze$^\star$} &  100 & $\textbf{4.45}$ & $\textbf{5.27}$ & $\textbf{4.94}$ & $\textbf{5.60}$ & $\textbf{5.07}$ \\
				\hline
			\end{tabular}}

	    \caption{Comparison with SOTA methods. Results are reported by angular error in degrees, bold and underline denotes the best and the second best result among each column on one specific task. $^\ddagger$ expresses the model employs ResNet-50 as  backbone, ${^*}$ indicates that experimental settings are different, $^\star$ denotes model is fine-tuned on target-domain.}
		\label{tab:full_comp}
	\end{table}	

    Quantitative result of four cross-domain gaze estimation tasks are shown in Tab. \ref{tab:full_comp}.
    The second row shows the comparison between our method and SOTA domain generalization (DG) methods.
    The CLIP-Gaze${^-}$ is the plain CLIP-Gaze framework without PCO module and feature rank loss. It improves the generalizable capablity of Baseline and achieves the comparable performance with Xu $\textit{et al.}$'s method using ResNet-18 as the backbone.
    The complete CLIP-Gaze achieves the best overall performance. It shows the state-of-the-art performance on three cross-domain evaluation tasks and achieves similar performance to the best method for $\mathcal{D}_\mathrm{E}$$\rightarrow$$\mathcal{D}_\mathrm{D}$, which proves the effectiveness of our proposed PCO module and feature rank loss.

    Besides, we provide the comparison results with SOTA unsupervised domain adaption (UDA) methods in the third row of Tab. \ref{tab:full_comp}. Note that UDA methods require a small number of unlabeled target domain samples. It can be observed that our method demonstrates advanced performance with no access to target domain data.
    Specifically, we surpass LatentGaze on task $\mathcal{D}_\mathrm{E}$$\rightarrow$$\mathcal{D}_\mathrm{D}$, achieve  better performance than Liu $\textit{et al.}$.’s work on task $\mathcal{D}_\mathrm{G}$$\rightarrow$$\mathcal{D}_\mathrm{M}$ and outperform PnP-GA and Liu $\textit{et al.}$.’s method on task $\mathcal{D}_\mathrm{G}$$\rightarrow$$\mathcal{D}_\mathrm{D}$.
    It demonstrates the strength of our proposed method since it does not need any target domain information.

    To further demonstrate the improvement of the proposed method, we randomly choose 100 target samples with labels for fine-tuning on our baseline and CLIP-Gaze model. The evaluation results in the last row of Tab. \ref{tab:full_comp} show that our fine-tuned model consistently outperforms the baseline after fine-tuning. The details of the fine-tuning experiments can be found in the supplementary materials.

	\begin{table}[ht]
		\renewcommand\arraystretch{1.5}
		\setlength{\tabcolsep}{1mm}
		\centering
		\scalebox{1}{
			\begin{tabular}{|l|c|c|c|c|c|}
				\hline
				\multirow{2}{*}{Methods}&
                $\mathcal{D}_\mathrm{E}  $ &
                $\mathcal{D}_\mathrm{E}  $ &
                $\mathcal{D}_\mathrm{G}  $ &
                $\mathcal{D}_\mathrm{G}  $ &
                \multirow{2}{*}{Avg}\\
                &
                $ \rightarrow \mathcal{D}_\mathrm{M}$ &
                $ \rightarrow \mathcal{D}_\mathrm{D}$ &
                $ \rightarrow \mathcal{D}_\mathrm{M}$ &
                $ \rightarrow \mathcal{D}_\mathrm{D}$ &
                \\
				\hline
				\hline
				Baseline & $8.35$ & $9.66$ & $7.58$ & $9.01$  & $8.65$ \\
				w/o TPT& $\underline{6.72}$ & $8.16$ & $\underline{7.07}$ & $7.64$  &  7.40\\ 	
				CoOp & $7.44$ & $\textbf{7.42}$ & $7.41$ & $\underline{7.15}$ &  $\underline{7.36}$\\
				CoCoOp & $7.56$ & $8.03$ & $7.52$ & $8.12$  &   7.81\\
				CoCoOp $^*$ & $7.36$ & $7.77$ & $7.31$ & $8.46$  &  7.73\\			
				PCO& $\textbf{6.41}$ & $\underline{7.51}$ & $\textbf{6.89}$ & $\textbf{7.06}$  & $\textbf{6.97}$ \\
				\hline
		\end{tabular}}
		\caption{Comparison of different text prompt tuning methods for gaze model cross-domain evaluation in four tasks. Bold indicates the best results in each column, and underline denote the second best result results in each column.}
		\label{tab:TPT_comp}
	\end{table}

	\begin{table}[ht]
		\renewcommand\arraystretch{1.5}
		\setlength{\tabcolsep}{1mm}
		\centering
		\scalebox{1}{
			\begin{tabular}{|l|c|c|c|c|c|}
				\hline
				\multirow{2}{*}{Methods}&
                $\mathcal{D}_\mathrm{E}  $ &
                $\mathcal{D}_\mathrm{E}  $ &
                $\mathcal{D}_\mathrm{G}  $ &
                $\mathcal{D}_\mathrm{G}  $ &
                \multirow{2}{*}{Avg}\\
                &
                $ \rightarrow \mathcal{D}_\mathrm{M}$ &
                $ \rightarrow \mathcal{D}_\mathrm{D}$ &
                $ \rightarrow \mathcal{D}_\mathrm{M}$ &
                $ \rightarrow \mathcal{D}_\mathrm{D}$ &
                \\
				\hline
				\hline
				Baseline & $8.35$ & $9.66$ & $7.58$ & $9.01$  & $8.65$ \\
				\hline
				Appearance($\mathcal{A}$)& $7.32$ & $\textbf{7.09}$ & $7.49$ & $7.54$  & $7.36$ \\		
				Wearable($\mathcal{W}$) & $7.33$ & $7.95$ & $7.57$ & $7.45$ & $7.58$ \\
				Quality($\mathcal{Q}$) & $7.30$ & $7.36$ & $7.24$ & $7.49$  & $7.35$ \\			
				\hline
				$\mathcal{W} +\mathcal{Q}$& $6.75$ & $7.87$ & $6.99$ & $7.95$  & $7.39$ \\		
				$\mathcal{A} +\mathcal{Q}$ & $6.91$ & $7.23$ & $\textbf{6.85}$ & $7.42$ & $7.10$ \\
				$\mathcal{A}+\mathcal{W}$ & $7.18$ & $7.32$ & $7.23$ & $7.66$  & $7.35$ \\
				\hline
				$\mathcal{A}+\mathcal{W}+\mathcal{Q}$& $\textbf{6.41}$ & $7.51$ & $6.89$ & $\textbf{7.06}$  & $\textbf{6.97}$ \\
				\hline
		\end{tabular}}
		\caption{Comparison of different gaze-irrelevant combinations for gaze model generalization in four cross-domain tasks. Bold indicates the best results in each column.}
		\label{tab:attributes}
	\end{table}

	\subsection{Ablation Study}
	\subsubsection{Ablation Study of Text Prompt Tuning methods}
        In this section, we compare different TPT methods in  Tab. \ref{tab:TPT_comp}, where “Baseline” is the same as in Tab. \ref{tab:full_comp}.
        We denote CLIP-Gaze without text prompt tuning as w/o TPT, which achieved a significant improvement over the Baseline.
        In our TPT experiments,
        CoOp\cite{zhou2022learning} learns only one prompt for each class, which only improves the performance slightly.
        CoCoOp\cite{zhou2022conditional} introduces instance-conditional text embedding for prompt tuning, but the instance containing gaze-relevant feature is detrimental to CLIP-Gaze, thus resulting in  worse performance. CoCoOp$^*$ only uses one user face image embedding as a conditional embedding for prompt tuning, but this image embedding may also contain impure identity features that may affect prompt tuning. Our PCO method achieves the best performance,
       which demonstrates using the 3DMM model to extract personalizing features and introducing the identity-conditional text embedding can learn more suitable prompts.

	\begin{table}[ht]
		\renewcommand\arraystretch{1.5}
		\setlength{\tabcolsep}{1mm}
		\centering
		\scalebox{0.97}{
			\begin{tabular}{|l|c|c|c|c|c|}
				\hline
				\multirow{2}{*}{Methods}&
                $\mathcal{D}_\mathrm{E}  $ &
                $\mathcal{D}_\mathrm{E}  $ &
                $\mathcal{D}_\mathrm{G}  $ &
                $\mathcal{D}_\mathrm{G}  $ &
                \multirow{2}{*}{Avg}\\
                &
                $ \rightarrow \mathcal{D}_\mathrm{M}$ &
                $ \rightarrow \mathcal{D}_\mathrm{D}$ &
                $ \rightarrow \mathcal{D}_\mathrm{M}$ &
                $ \rightarrow \mathcal{D}_\mathrm{D}$ &
                \\
				\hline
				\hline
				Baseline & 8.35 & 9.66 & 7.58 & 9.01  & 8.65 \\
				\hline
				$+\mathcal{L}_{d}$ & 7.82 & 9.32 & 7.24 & 9.14 & 8.38 \\		
				$+\mathcal{L}_{d}+\mathcal{L}_{ir}$ & $\underline{6.54}$ & 7.94 & 7.44 & 7.57 & 7.37\\
				\hline
				$+\mathcal{L}_{d}+\mathcal{L}_{ir}+\mathcal{L}_{CR}$ & 6.96 & 7.89 & 7.24 & 7.28 & 7.34 \\
				$+\mathcal{L}_{d}+\mathcal{L}_{ir}+\mathcal{L}_{re}^{L1}$ & 6.81 & 7.91 & 7.07 & $\textbf{6.84}$ & 7.16 \\			
				$+\mathcal{L}_{d}+\mathcal{L}_{ir}+\mathcal{L}_{re}^{L2}$ & 6.79 & $\underline{7.59}$ & $\underline{6.92}$ & 7.21 & $\underline{7.13}$ \\
				$+\mathcal{L}_{d}+\mathcal{L}_{ir}+\mathcal{L}_{re}^{KL}$ & 6.96 & 7.89 & 6.97 & 7.08 & 7.23 \\
				$+\mathcal{L}_{d}+\mathcal{L}_{ir}+\mathcal{L}_{re}$ & $\textbf{6.41}$ & $\textbf{7.51}$ & $\textbf{6.89}$ & $\underline{7.06}$ & $\textbf{6.97}$ \\
				\hline
		\end{tabular}}
		\caption{Ablation study on loss functions. Results are reported by angular error in degrees.}
		\label{tab:loss_func}
	\end{table}

	\subsubsection{Ablation Study of Attributes and Conditions}
        To investigate the effects of different gaze-irrelevant factors on the gaze model, we conduct different factor combinations experiments, as shown in Tab. \ref{tab:attributes}. We can draw the following three conclusions: (1) The combination of multiple group of gaze-irrelevant factors outperforms the single group factors individually, possibly because filtering out more gaze-irrelevant features can make the model more generalizable; (2) Appearance($\mathcal{A}$) and image quality($\mathcal{Q}$) have larger impacts on the generalization ability of the gaze model than other factors, possibly because they account for the major domain gap; (3) The full combination of all factors($\mathcal{A}+\mathcal{W}+\mathcal{Q}$) achieves the best performance.

	\subsubsection{Ablation Study of Loss Functions}
        Compared with our baseline, we propose three loss functions to achieve the goal of extracting robust gaze-relevant features for gaze estimation.
        To investigate the effectiveness of $\mathcal{L}_{d}$, $\mathcal{L}_{ir}$ and $\mathcal{L}_{re}$, we conduct experiments to prove their effects by gradually increasing the loss term in baseline model training.
        For $\mathcal{L}_{re}$, we compare the CRLoss $\mathcal{L}_{CR}$ and our proposed rank loss and its variants, such as$\mathcal{L}_{re}^{L1}$, $\mathcal{L}_{re}^{L2}$, and $\mathcal{L}_{re}^{KL}$, which compute the L1, L2, and KL losses between the feature similarity and label similarity sequences of sample pairs, respectively.

        Based on the results shown in in the third row of Tab.~\ref{tab:loss_func}, we can see distilling gaze features to CLIP feature space enhances the average cross-domain performance, and there is a significant performance improvement  after eliminating gaze-irrelevant features, this proves our proposed framework is efficient and superior.
        On this basis, we compare different loss forms of the relationship among sample features, it can be observed in the last row, the improvement brought by $\mathcal{L}_{CR}$ is slight and even worse than these variants of $\mathcal{L}_{re}$ that directly align the feature similarity to label similarity. Instead of learning the absolute values of feature similarities about sample pairs, our $\mathcal{L}_{re}$ constrains their relative magnitudes and achieve the best overall performance, these results above demonstrate the effectiveness of our framework and proposed loss functions.

    \begin{figure}[ht]
         \centering
         \includegraphics[width=0.43\textwidth]{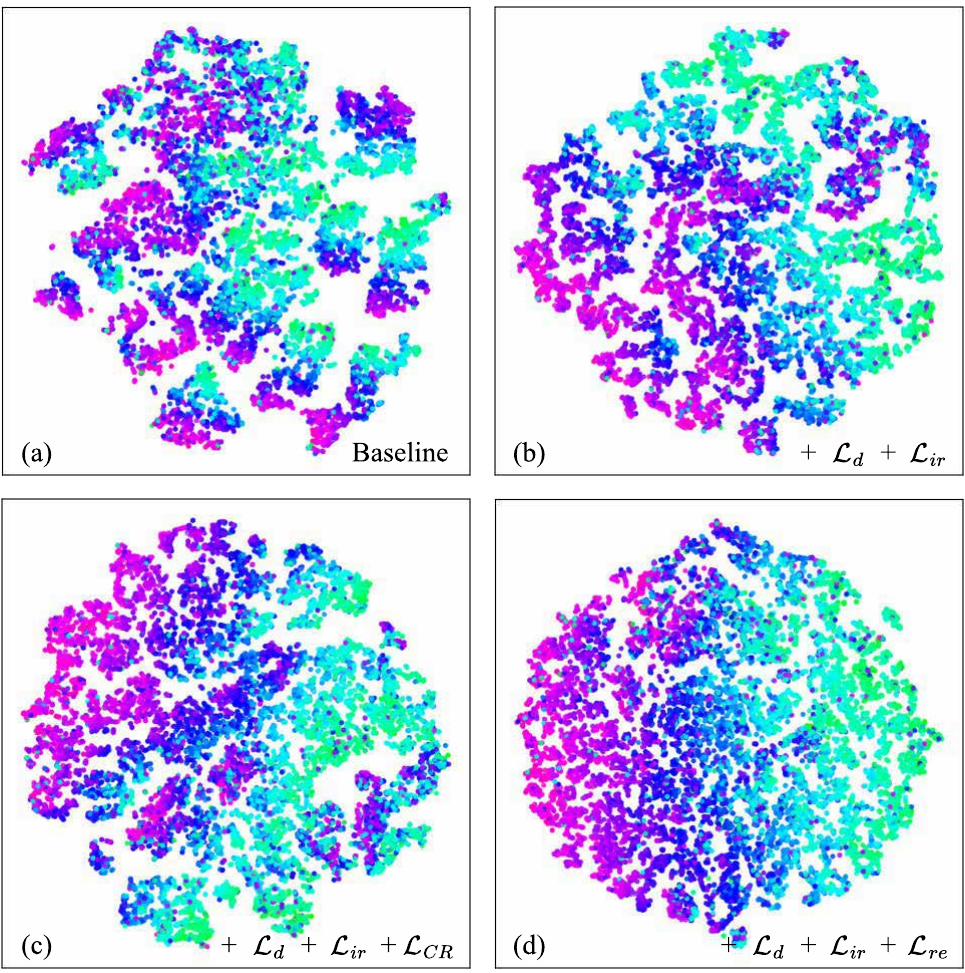}
         \caption{Visualization of the feature distribution. Different colors denotes different gaze directions and close gaze directions share similar colors. (Best viewed in color).}
         \label{tag:vis}
    \end{figure}

	\subsection{Visualization of Extracted Features}
	To compare and analyze the extracted features $\boldsymbol{f}^{re}$ of different models, we follow the manner in \cite{cdg} to visualize the distribution of feature on the task $\mathcal{D}_\mathrm{G}$$\rightarrow$$\mathcal{D}_\mathrm{D}$ with t-SNE\cite{tsne}.
    Fig. \ref{tag:vis} displays the visualization results from four different models in Tab. \ref{tab:loss_func}, where feature points with similar gaze directions share similar colors.

    For the Baseline model, the features with different gaze directions are mixed together and the feature cluster is quite dispersed which is not sensible for regression task.
    After eliminating the gaze-irrelevant parts from extracted feature, the model of Fig. \ref{tag:vis} (b) which omit the $\mathcal{L}_{re}$ in CLIP-Gaze shows the overall feature distribution becoming ordered on gaze directions, and the similar colors are close in the feature space.
    However, there is an unreasonably purple features cluster appears in the green area in the upper right of the box.
    Similarly, as shown in Fig. \ref{tag:vis} (c), the model with additional $\mathcal{L}_{CR}$ does not improve the feature distribution compared with the model of Fig. \ref{tag:vis} (b). This confirms that simply pushing features away or pulling features together by a fixed threshold is not optimal.
    In general, CLIP-Gaze has the most reasonable feature distribution and the visualization is shown in Fig. \ref{tag:vis} (d), the gaze direction similarities and feature similarities have a strong correlation, this means our proposed feature rank loss $\mathcal{L}_{re}$ is effective.
    More visualizations provided in the supplementary materials.

    \section{Conclusion}
	In this paper, we propose a domain-generalization framework for gaze estimation models, which leverages visual-linguistic models to handle diverse target domains. Specifically, we define the gaze-irrelevant factors, such as face appearance, wearable, and image quality, and construct gaze-irrelevant features using language descriptions. Then, we decouple the gaze features from the CLIP feature space to enhance the model’s generalization ability. To enchance the performance, a personalized context optimization method is proposed for text prompt tuning, and a rank loss is designed to  learn a more reasonable gaze feature distribution. Our proposed framework achieves state-of-the-art performance on domain generalization for gaze estimation tasks.

\bibliography{aaai24}

\end{document}